\documentclass[a4paper,fleqn]{cas-dc}
\usepackage[authoryear,longnamesfirst]{natbib}
\usepackage{graphicx}
\usepackage[lofdepth,lotdepth]{subfig}
\usepackage{array}
\def\tsc#1{\csdef{#1}{\textsc{\lowercase{#1}}\xspace}}
\tsc{WGM}
\tsc{QE}
\begin{document}
\let\WriteBookmarks\relax
\def\floatpagepagefraction{1}
\def\textpagefraction{.001}

\shorttitle{Geographic Fake Image Detection}

\shortauthors{Hong-Shuo Chen et al.}

\title [mode = title]{Geo-DefakeHop: High-Performance Geographic Fake Image Detection}  

\author[1]{Hong-Shuo Chen}[orcid=0000-0003-1116-939X]
\cormark[1]
\ead{hongshuo@usc.edu}
\author[1]{Kaitai Zhang}
\author[2]{Shuowen Hu}
\author[2]{Suya You}
\author[1]{C.-C. Jay Kuo}

\affiliation[1]{organization={University of Southern California},
            city={Los Angeles},
            state={California},
            country={USA}}
            
\affiliation[2]{organization={Army Research Laboratory},
            city={Adelphi},
            state={Maryland},
            country={USA}}
\cortext[1]{Corresponding author}

\begin{abstract}
A robust fake satellite image detection method, called Geo-DefakeHop, is
proposed in this work.  Geo-DefakeHop is developed based on the parallel
subspace learning (PSL) methodology.  PSL maps the input image space
into several feature subspaces using multiple filter banks.  By
exploring response differences of different channels between real and
fake images for a filter bank, Geo-DefakeHop learns the most
discriminant channels and uses their soft decision scores as features.
Then, Geo-DefakeHop selects a few discriminant features from each filter
bank and ensemble them to make a final binary decision.  Geo-DefakeHop
offers a light-weight high-performance solution to fake satellite images
detection. Its model size is analyzed, which ranges from 0.8 to 62K
parameters. Furthermore, it is shown by experimental results that it
achieves an F1-score higher than 95\% under various common image
manipulations such as resizing, compression and noise corruption. 
\end{abstract}


\maketitle

\section{Introduction}\label{sec:intro}

Artificial intelligence (AI) and deep learning (DL) techniques have made
significant advances in recent years by leveraging more powerful
computing resources and larger collected and labeled datasets, 
creating breakthroughs in various fields such as computer vision,
natural language processing, and robotics. Geospatial science
\citep{janowicz2020geoai} and remote sensing \citep{ma2019deep} also
benefit from this development, involving increased application of AI to process
data arising from cartography and geographic information science (GIS)
more effectively.  Despite the countless advantages brought by AI,
misinformation over the Internet, ranging from fake news
\citep{radford2019language} to fake images and videos
\citep{dolhansky2020deepfake, zi2020wilddeepfake,li2020celeb,
zhou2021face, rossler2018faceforensics, jiang2020deeperforensics}, poses
a serious threat to our society.  It is important to be able to judge the
authenticity of online content \citep{fbi2019}. 

In the context of geospatial science, satellite images are utilized in
various applications such as weather prediction, agriculture crops
prediction, flood and fire control. If one cannot determine whether a
satellite image is real or fake, it would be risky to use it for
decision making. Fake satellite images have impacts on national security
as well.  For example, adversaries can create fake satellite images to hide
important military infrastructure and/or create fake ones to deceive
others.  Though government analysts could verify the authenticity of 
geospatial imagery leveraging other satellites or data sources, this would
be prohibitively time intensive. For the public, it would be extremely difficult
to verify the authenticity of satellite images.

It is becoming easier and easier to generate
realistically looking images due to the rapid growth of generative
adversarial networks (GANs) \citep{brock2018large, karras2019style,
park2019semantic, choi2018stargan, karras2017progressive,
zhu2017unpaired}. Typically, a base map of an input satellite image can be
first produced by one GAN. Then, a fake satellite image can be
generated by another GAN based on the base map.  Since generated
satellite images are difficult to discern by human eyes, there is an urgent
need to develop an automatic detection system that can find fake
satellite images accurately and efficiently. 

Little research has been done on fake satellite images detection due
to the lack of a proper fake satellite image dataset.  The first fake
satellite image dataset was recently proposed by Zhao {\em et al.}
\citep{zhao2021deep}. Furthermore, handcrafted spatial, histogram, and
frequency features were derived and a support vector machine (SVM)
classifier was trained to determine whether a satellite tile is real and
fake. It has an F1-score of 83\% in detection performance. We are not
aware of any DL solution to this problem yet but expect to see some
in the future. There are DL-based fake image detection methods for other
types of images. They will be reviewed in Sec. \ref{sec:review}.

A robust fake satellite image detection method, called Geo-DefakeHop, is
proposed in this work. Our research is based on one observation and
one assumption. The observation is that the human visual system (HVS)
\citep{hall1977nonlinear} has its limitation. That is, it behaves like a
low-pass filter and, as a result, it has a poor discriminant power for
high-frequency responses. The assumption is that GANs can generate
realistic images by reproducing low-frequency responses of synthesized
images well. Yet, it might not be able to synthesize both low and high-frequency components well due to its limited model complexity. If 
this assumption holds, we can focus on differences between higher frequency
components in differentiating true and fake images.  

This high-level idea can be implemented by a set of filters operating at all pixel
locations in parallel, known as a filter bank in signal processing. Each
filter offers responses of a particular frequency channel in the spatial
domain and these responses can be used to check the discriminant power
of a channel from the training data. To make the detection model more
robust, we adopt multiple filter banks, find discriminant channels from
each, and ensemble their responses to get the final binary decision.
Since multiple filter banks are used simultaneously, it is named
parallel subspace learning (PSL).  The proposed Geo-DefakeHop offers a
lightweight, high-performance and robust solution to fake satellite
images detection. Its model ranges from 0.8 to 62K parameters. It
achieves an F1-score higher than 95\% under various common image
manipulations such as resizing, compression and noise corruption. 

The contributions of this work are summarized below.
\begin{enumerate}
\item We propose a fake satellite image detection method, called
Geo-DefakeHop, which exploits the PSL methodology to extract
discriminant features with the implementation of multiple filter banks. 
\item We use the heat map to visualize prediction results and spatial
responses of different frequency channels for Geo-DefakeHop's
interpretability. 
\item We conduct extensive experiments to demonstrate the high
performance of Geo-DefakeHop and its robustness against various image
manipulations. 
\end{enumerate}
The rest of this paper is organized as follows.  Related work is
reviewed in Sec. \ref{sec:review}. The Geo-DefakeHop method is presented
in Sec. \ref{sec:method}. Experiments are shown in Sec.
\ref{sec:experiments}. Finally, concluding remarks are given in Sec.
\ref{sec:conclusion}. 

\section{Related Work}\label{sec:review}

\subsection{Fake images generation}

\begin{figure}[!t]
\centering
\includegraphics[width=0.9\linewidth]{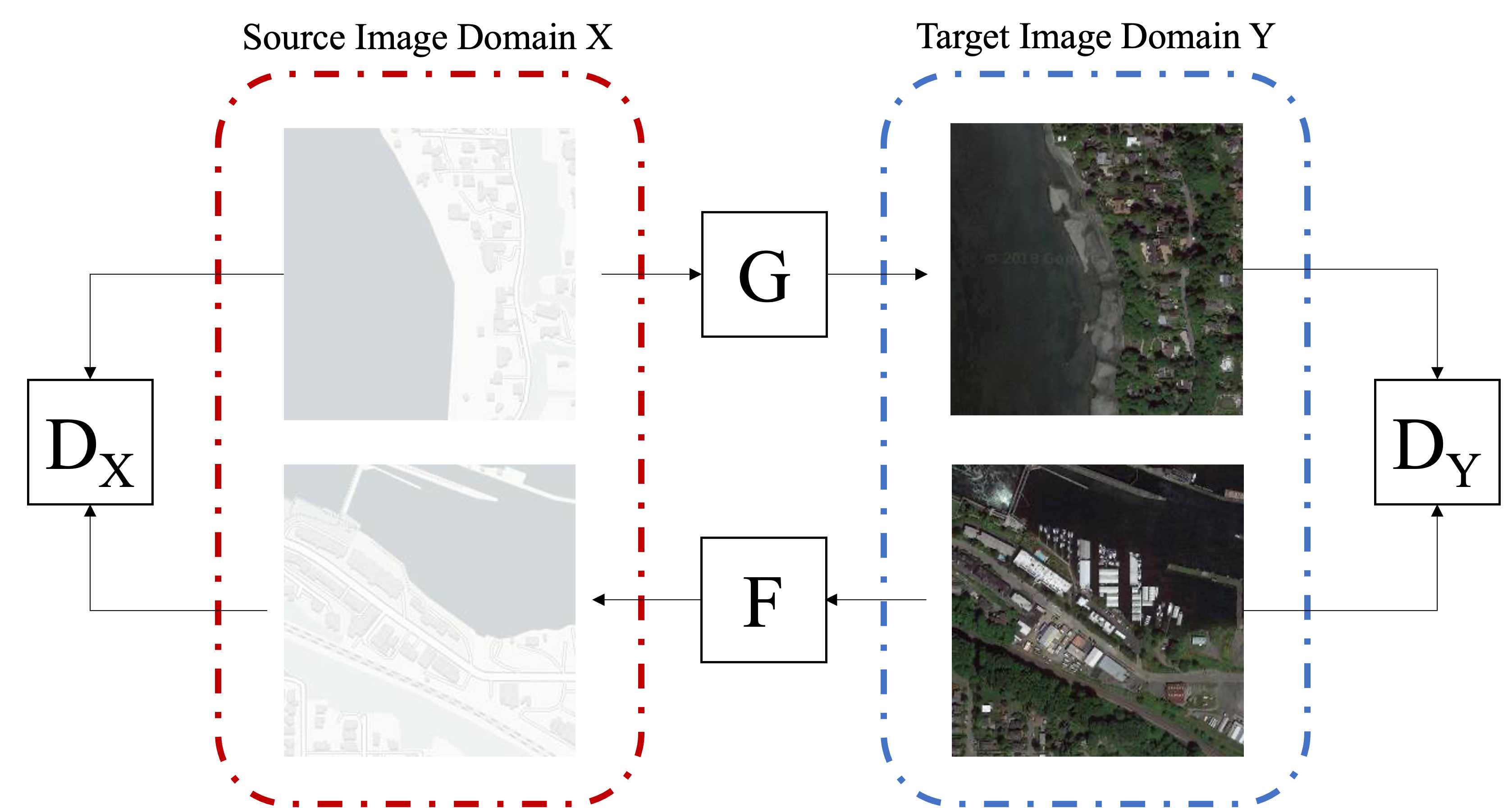}
\caption{Illustration of CycleGAN and its application to fake 
satellite image generation. }\label{fig:fig_pre}
\end{figure}

GANs provide powerful machine learning models for image-to-image
translation.  It consists of two neural networks in the training
process: a generator and a discriminator. The generator attempts to
generate fake images to fool the discriminator while the discriminator
tries to distinguish generated fake images from real ones. They are
jointly trained via end-to-end optimization with an adversarial loss.
In the inference stage, only the generator is needed.  For
image-to-image translation, GANs need paired images to train, say, image
$A$ from source domain $X$ and image $B$ from target domain $Y$, where
images $A$ and $B$ describe the same object or scene.  This limits GAN's
power since paired images are challenging to collect.  Cycle-consistent
GAN (CycleGAN) \citep{zhu2017unpaired} was proposed to overcome this
difficulty.  It learns to translate an image from source domain $X$ to
target domain $Y$ without paired images.  To achieve this, CycleGAN
supplements the desired mapping, $G$, with an inverse mapping, $F$,
which maps images in domain $Y$ back to domain $X$. It imposes a cycle
consistency loss to minimize the distance between $F(G(X))$ and $X$.
CycleGAN has been applied to many applications, including fake satellite
image generation \citep{zhao2021deep} as shown in Fig.
\ref{fig:fig_pre}, where a base map of an input satellite image is first
generated by one GAN. Then, a fake satellite image is generated by
another GAN based on the base map. 

\subsection{Fake images detection}

Most fake image detection methods adopt convolution neural networks
(CNNs).  \citep{wang2020cnn} used the real and fake images generated by
ProGAN \citep{karras2017progressive} as the input of ResNet-50
pretrained by the ImageNet. \citep{zhang2019detecting} generated fake
images with their designed GAN, called AutoGAN, and claimed that CNN
trained by their simulated images could learn artifacts of fake images.
\citep{nataraj2019detecting} borrowed the idea from image steganalysis
and used the co-occurrence matrix as input to a customized CNN so that
it can learn the differences between real and fake images.  By following
this idea, \citep{barni2020cnn} added the cross-band co-occurrence
matrix to the input so as to increase the stability of the model.
\citep{guarnera2020deepfake} utilized the EM algorithm and the KNN
classifier to learn the convolution traces of artifacts generated by
GANs.  Little research has been done to date on fake satellite images detection
due to the lack of available datasets. \citep{zhao2021deep} proposed the
first fake satellite image dataset with simulated satellite images from
three cities (i.e., Tacoma, Seattle and Beijing). Furthermore, it used
26 hand-crafted features to train an SVM classifier for fake satellite
image detection.  The features can be categorized into spatial,
histogram and frequency three classes. Features of different classes are
concatenated for performance evaluation. In Sec. \ref{sec:experiments},
we will benchmark our proposed Geo-DefakeHop method with the method in
\citep{zhao2021deep}. 

\subsection{PixelHop and Saab Transform}\label{subsec:pixelhop}

\begin{figure*}[h]
\centering
\includegraphics[width=1\textwidth]{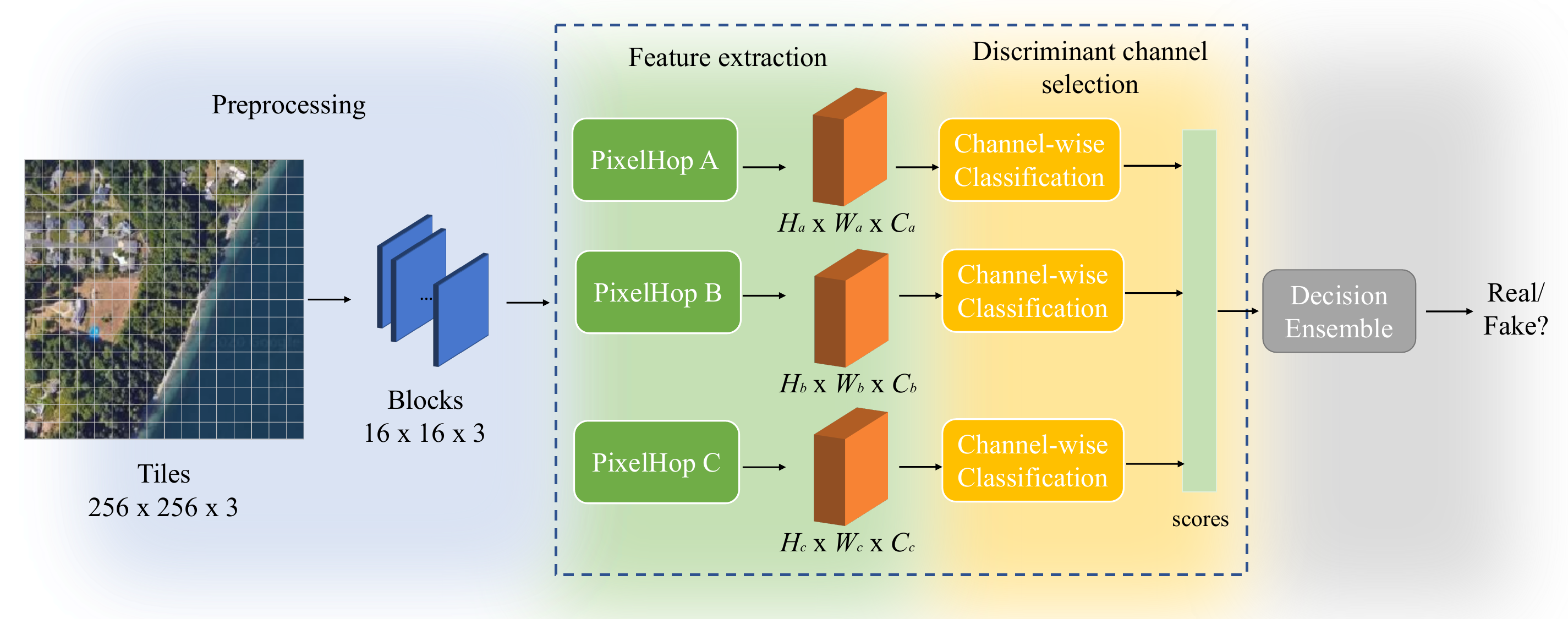}
\caption{An overview of the Geo-DefakeHop method, where the input is an
image tile and the output is a binary decision on whether the input
is an authentic or a fake one. First, each input title is partitioned into
non-overlapping blocks of dimension $16 \times 16 \times 3$.  Second, each
block goes through one PixelHop or multiple PixelHops, each
of which yields 3D tensor responses of dimension $H \times W \times
C$. Third, for each PixelHop, an XGBoost classifier is applied
to spatial samples of each channel to generate channel-wise (c/w) soft
decision scores and a set of discriminant channels are selected
accordingly. Last, all block decision scores are ensembled to generate the
final decision of the image tile.}\label{fig:fig_geo}
\end{figure*}

The PixelHop concept was introduced by Chen {\em et al.} in
\citep{chen2020pixelhop}.  Each PixelHop has local patches of the same
size as its input.  Suppose that local patches are of dimension $L=s_1
\times s_2 \times c$, where $s_1\times s_2$ is the spatial dimension and
$c$ is the spectral dimension. A PixelHop defines a mapping from pixel
values in a patch to a set of spectral coefficients, which is called the
Saab transform \citep{kuo2019interpretable}.  The Saab transform is a
variant of the principal component analysis (PCA).  For standard PCA, we
subtract the ensemble mean and then conduct eigen-analysis on the
covariance matrix of input vectors.  The ensemble mean is difficult to
estimate if the sample size is small.  The Saab transform decomposes the
$n$-dimensional signal space into a one-dimensional DC (direct current)
subspace and an $(n-1)$-dimensional AC (alternating current) subspace.
Signals in the AC subspace have an ensemble mean close to zero.  Then, we
can apply PCA to the AC signal and decompose it into $(n-1)$ channels.
Saab coefficients are unsupervised data-driven features since Saab
filters are derived from the local correlation structure of pixels. 

\subsection{Differences between DefakeHop and Geo-DefakeHop}

The Saab transform can be implemented conveniently with filter banks. It
has been successfully applied to many application domains. Examples
include~\citep{chen2021defakehop, zhang2021anomalyhop, liu2021voxelhop,
zhang2020pointhop}.  Among them, DefakeHop \citep{chen2021defakehop} is
closest to our current work.  However, there are three main differences
between DefakeHop and Geo-DefakeHop proposed here.  First, DefakeHop was
proposed to detect deepfake videos, where the main target is human
faces.  Here, our target is to detect fake satellite images.  Since the
targets are different, we need to tailor the current work accordingly.
Second, DefakeHop used multi-stage cascaded PixelHop units to extract
features from human eyes, nose and mouth regions. It focused on low-frequency channels and discarded high-frequency channels. In contrast,
we abandon multi-stage Saab transforms in cascade, and adopt multiple
one-stage Saab transforms that operate in parallel.  We show that high-frequency channels are more discriminant than low-frequency channels for
fake image detection.  Third, no image manipulation was tested in
DefakeHop. The robustness issue is carefully examined in the current
work.  We compare the performance of DefakeHop and Geo-DefakeHop in Sec.
\ref{sec:experiments} and show that Geo-DefakeHop outperforms DefakeHop
by a significant margin. 

\section{Geo-DefakeHop Method}\label{sec:method}

Our idea is motivated by the observation that GANs fail to generate
high-frequency components such as edges and complex textures well.  It
is pointed out by \citep{frank2020leveraging} that GANs have
inconsistencies between the spectrum of real and fake images in
high-frequency bands. Another evidence is that images generated by
simple GANs are blurred and unclear. Blurry artifacts are reduced and
more details are added by advanced GANs to yield higher quality fake
images. Although these high quality simulated images look real to human
eyes because of the limitation of the HVS, it does not mean that the
high-frequency fidelity loss is not detectable by machines.  Another
shortcoming of generated images is periodic patterns introduced by
convolution and deconvolution operations in GAN models as reported in
\citep{guarnera2020deepfake}. GANs often use a certain size of
convolution and deconvolution filters (e.g., $3\times3$ or $5\times5$).
They leave traces on simulated images in form of periodic patterns in
some particular frequency bands. Sometimes, when GAN models do not
perform well, they can be observed by human eyes. 

Being motivated by the above two observations, we proposed a new method
for fake satellite image detection as shown in Fig. \ref{fig:fig_geo}.
It consists of four modules:
\begin{enumerate}
\item Preprocessing: Input image tiles are cropped into non-overlapping
blocks of a fix size. 
\item Joint spatial/spectral feature extraction via PixelHop:  The
PixelHop has a local patch as its input and applies a set of Saab
filters to pixels of the patch to yield a set of joint spatial/spectral
responses as features for each block.
\item Channel-wise classification, discriminant channels selection and
block-level decision ensemble: We apply an XGBoost classifier to spatial
responses of each channel to yield a soft decision, and select
discriminant channels accordingly. Then, the soft decisions from
discriminant channels of a single PixelHop or multiple PixelHops are
ensembled to yield the block-level soft decision.
\item Image-level decision ensemble: Block-level soft decisions 
are ensembled to yield the image-level decision. 
\end{enumerate}
They are elaborated below.

\begin{figure*}[ht]
\centering
\subfloat[][Without perturbation]{
\includegraphics[width=0.45\textwidth]{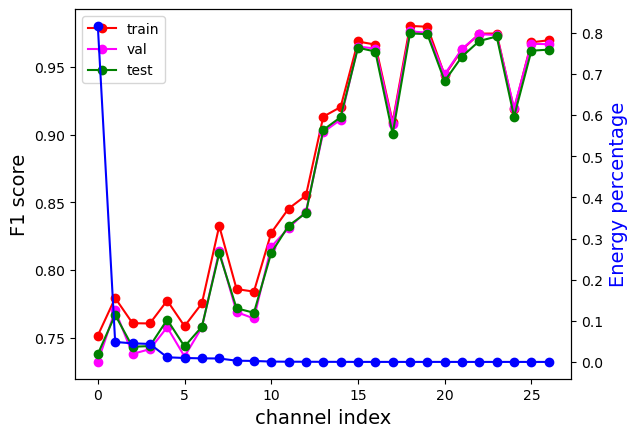}
\label{fig:ori}}
\qquad
\subfloat[][Resizing to $64 \times 64$]{
\includegraphics[width=0.45\textwidth]{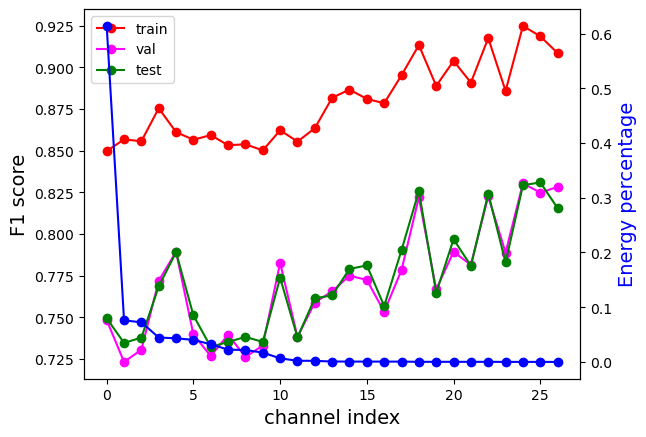}
\label{fig:resize}}
\qquad
\subfloat[][Adding Gaussian noise with $\sigma = 0.1$]{
\includegraphics[width=0.45\textwidth]{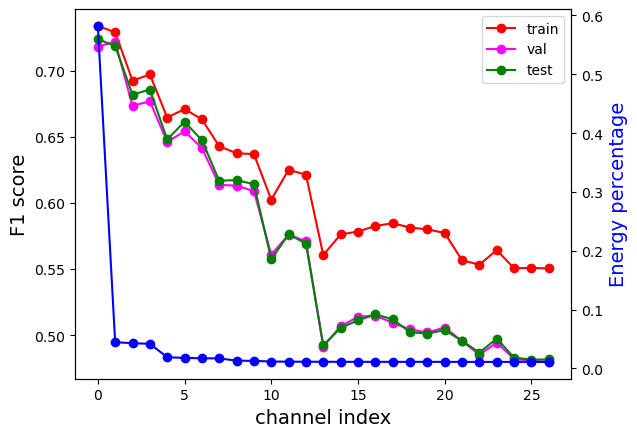}
\label{fig:noise}}
\qquad
\subfloat[][JPEG compression with $Q = 75$]{
\includegraphics[width=0.45\textwidth]{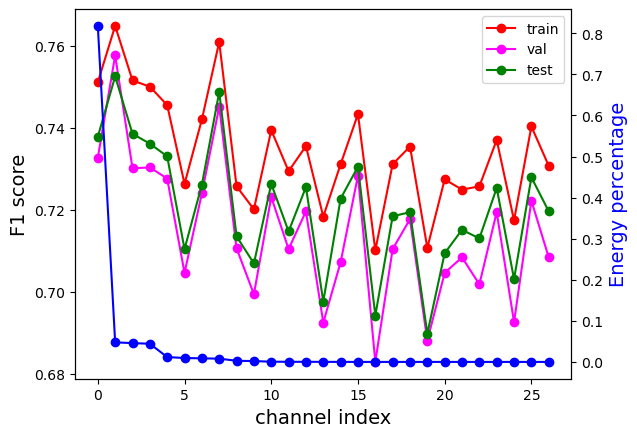}
\label{fig:jpeg}}
\caption{The channel-wise performance of four settings:
a) without perturbation, b) resizing, c) adding Gaussian noise, and d)
JPEG compression. The channel 0 is DC (Direct Current) and from the
first channel to the 26th channel are corresponding to AC1 to AC26
(Alternating Current). The blue line is the energy percentage of each
channel and the red, magenta and green lines are the F1-score of the
training, validation and testing dataset. We observe that high-frequency
channels without perturbation in \ref{fig:ori} has a higher performance.
After applying resizing, adding Gaussian noise and compression, the
performance of high-frequency channels degrades as shown in
\ref{fig:resize}, \ref{fig:noise}, \ref{fig:jpeg}. The test score and
validation score are closely related, indicating that the validation score
can be used to select the discriminant channels.} \label{fig:cp}
\end{figure*}

\subsection{Preprocessing}

A color satellite image tile of spatial resolution $256 \times 256$
covers an area of one kilometer square as shown in the left of Figure
\ref{fig:fig_geo}. It is cropped into 256 non-overlapping blocks of
dimension $16 \times 16 \times 3$, where the last number 3 denotes the
R, G, B three color channels. Each block has homogeneous content such as
trees, buildings, land and ocean. 

\subsection{Joint spatial/spectral feature extraction via PixelHop}

As described in Sec. \ref{subsec:pixelhop}, a PixelHop has a local patch
of dimension $L=s_1 \times s_2 \times c$ as its input, where $s_1$ and
$s_2$ are spatial dimensions and $c$ is the spectral dimension. For
square patches, we have $s_1=s_2=s$. We set $s$ to 2, 3, 4 in the
experiments. Since the input has R, G, B three channels, $c=3$. 

The PixelHop applies $L$ Saab filters to pixels in the local patch,
including one DC filter and $(L-1)$ AC filters, to generate $L$
responses per patch. The AC filters are obtained via eigen-analysis of
AC components. The mapping from $L$ pixel values to $L$ filter responses
defines the Saab transform. Since the AC filters are derived from the
statistics of the input, the Saab transform is a data-driven transform. 

We adopt overlapping patches with stride equal to one. Then, for a block
of spatial size $16\times 16$, we obtain $W \times H$ patches, where
$W=17-s_1$ and $H=17-s_2$. As a result, the block output is a set of
joint spatial/spectral responses of dimension $W\times H \times L$.  To
give an example, if the local patch size is $3\times3\times3=27$, the
block output is a 3D tensor of dimension $14 \times 14 \times 27$. They
are used as features to be fed to the classifier in the next stage. 

\subsection{Channel-wise classification, discriminant channels selection
and block-level decision ensemble}

For each channel in a block, we have one response from each local patch
so that there are $W \times H$ responses in total. These responses form
a feature vector, and samples from blocks of training real/fake images
are used to train a classifier, leading to channel-wise classification.
The classifier can be any one used in machine learning such as Random
Forest, SVM, and XGBoost. In our experiments, the XGBoost classifier
\citep{chen2016xgboost} is chosen for its high performance.  XGBoost is
a gradient-boosting decision tree algorithm that can learn a nonlinear
data distribution efficiently. 

To evaluate the discriminant power of a channel, we divide the training
data into two disjoint groups: 1) data used to train the classifier, and
2) data used to validate the channel performance. The latter provides a
soft decision score predicted by the channel-wise classifier.  The
channel-wise performance evaluation reflects the generation power of a
GAN in various frequency bands. Some channels are more discriminant than
others because of the poor generation power of the GAN in that frequency
band. Selection of discriminant channels is based on the performance of
the validation data. 

\begin{table*}[ht]
\caption{Visualization of real and fake satellite images with heat maps,
where cold and warm colors indicate a higher probability of being real
and fake in the corresponding location, respectively.}\label{tab:example}
\begin{center}
\begin{tabular}{cccc}\hline\hline
 \multicolumn{2}{c}{Real} & \multicolumn{2}{c}{Fake} \\ \hline
 Input image &  Heat map & Input image &  Heat map \\ \hline
 \includegraphics[width=0.2\textwidth]{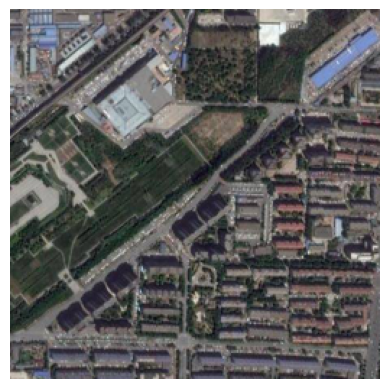} 
& \includegraphics[width=0.2\textwidth]{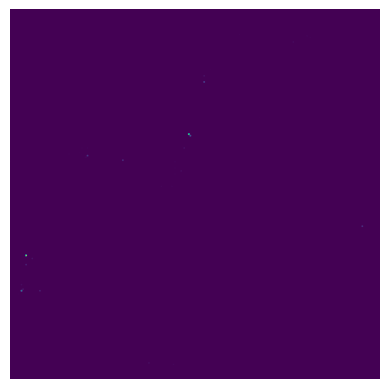} 
& \includegraphics[width=0.2\textwidth]{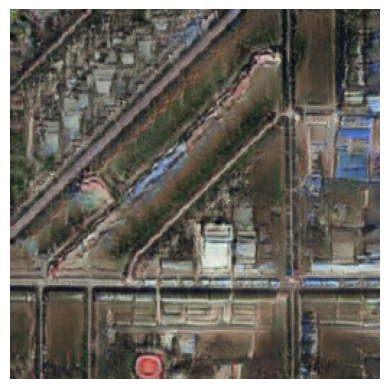} 
& \includegraphics[width=0.2\textwidth]{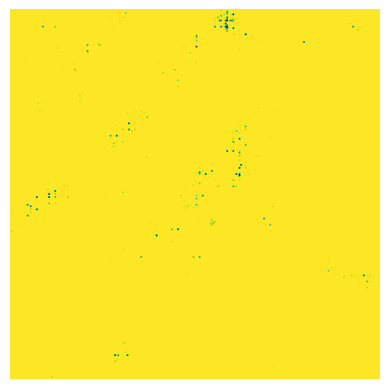}\\\hline
  \includegraphics[width=0.2\textwidth]{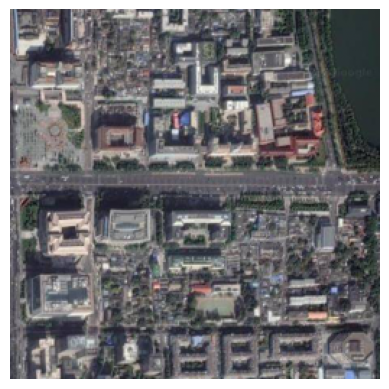} 
& \includegraphics[width=0.2\textwidth]{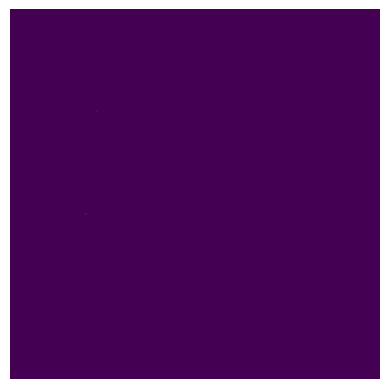} 
& \includegraphics[width=0.2\textwidth]{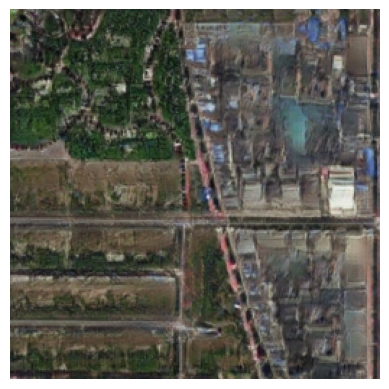} 
& \includegraphics[width=0.2\textwidth]{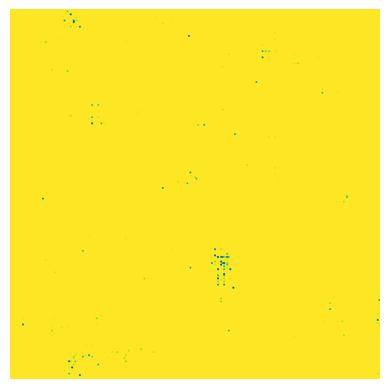}\\\hline
  \includegraphics[width=0.2\textwidth]{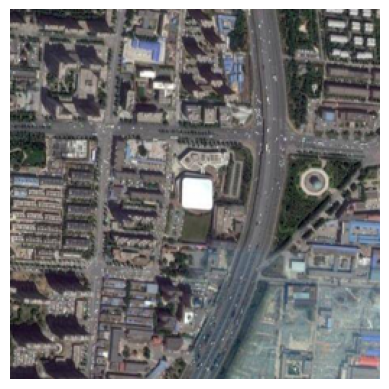} 
& \includegraphics[width=0.2\textwidth]{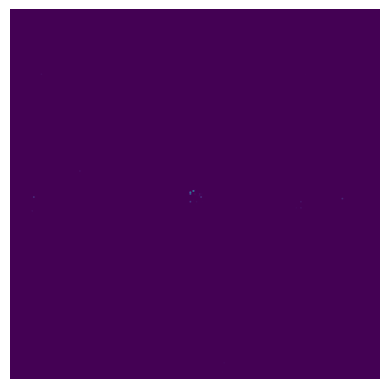} 
& \includegraphics[width=0.2\textwidth]{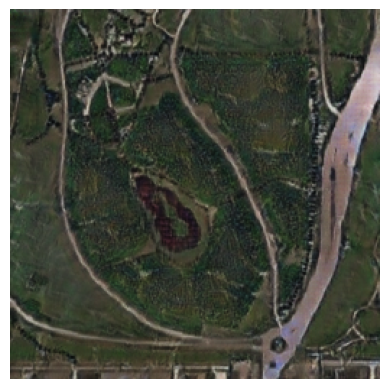} 
& \includegraphics[width=0.2\textwidth]{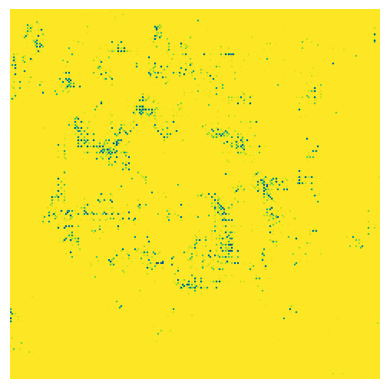}\\\hline
  \includegraphics[width=0.2\textwidth]{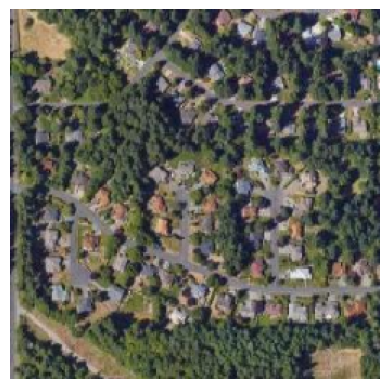} 
& \includegraphics[width=0.2\textwidth]{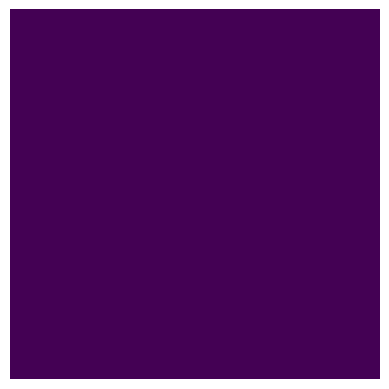} 
& \includegraphics[width=0.2\textwidth]{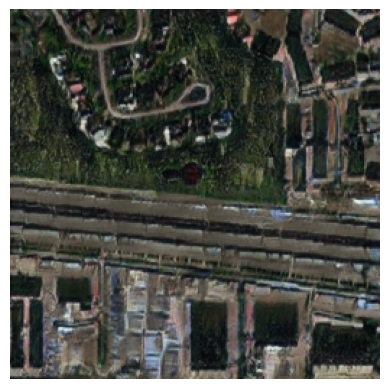} 
& \includegraphics[width=0.2\textwidth]{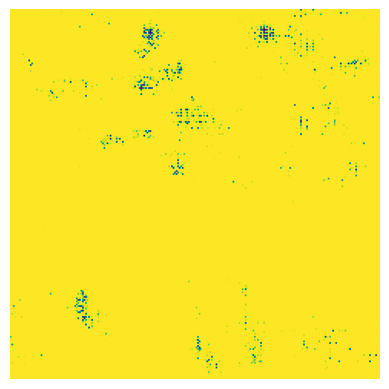}\\\hline
\end{tabular}
\end{center}
\end{table*}


\begin{table*}[ht]
\caption{Visualization of absolute values of Saab filter responses and
the detection heat maps for DC, AC1, AC11 and AC26 four channels, where
DC and AC1 are low-frequency channels, AC11 is a mid-frequency channel,
and AC26 is a high-frequency channel.  Cold and warm colors in heat maps
indicate a higher probability of being real and fake in the
corresponding location, respectively.}\label{tab:features}
\begin{center}
\begin{tabular}{c|c|c|c|c}\hline\hline
Index &  Name & Input image  & Saab features &  Heat map \\ \hline
0 & DC & \includegraphics[width=0.23\textwidth]{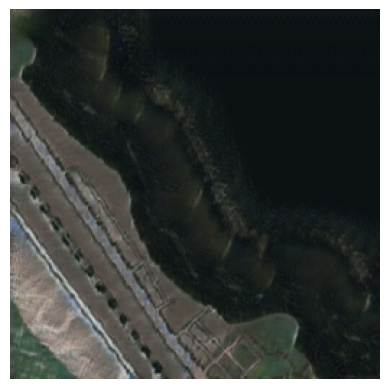} 
& \includegraphics[width=0.23\textwidth]{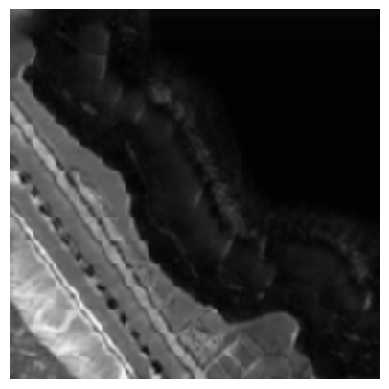} 
& \includegraphics[width=0.23\textwidth]{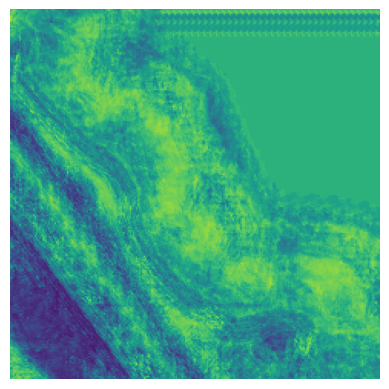} \\\hline
1 & AC1 & \includegraphics[width=0.23\textwidth]{figures/img_760.png} 
& \includegraphics[width=0.23\textwidth]{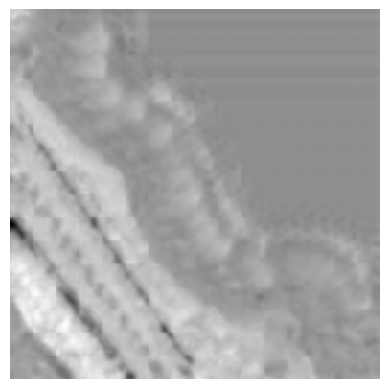} 
& \includegraphics[width=0.23\textwidth]{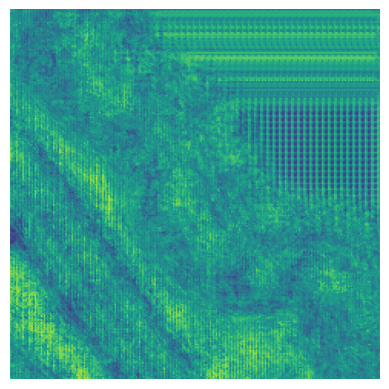} \\\hline
11 & AC11 & \includegraphics[width=0.23\textwidth]{figures/img_760.png} 
& \includegraphics[width=0.23\textwidth]{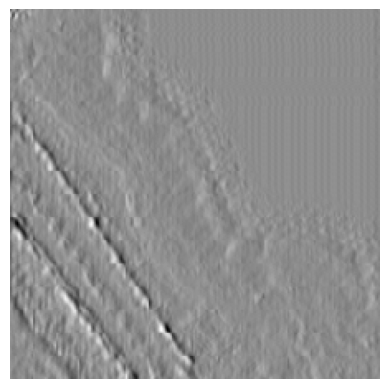} 
& \includegraphics[width=0.23\textwidth]{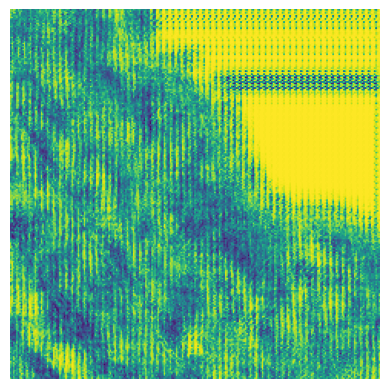} \\\hline
26 & AC26 & \includegraphics[width=0.23\textwidth]{figures/img_760.png} 
& \includegraphics[width=0.23\textwidth]{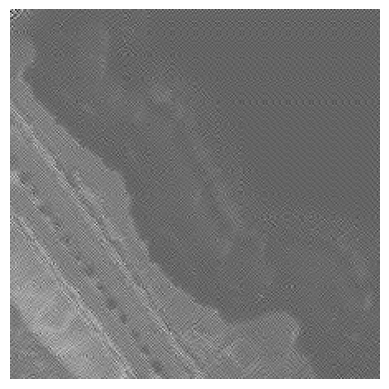} 
& \includegraphics[width=0.23\textwidth]{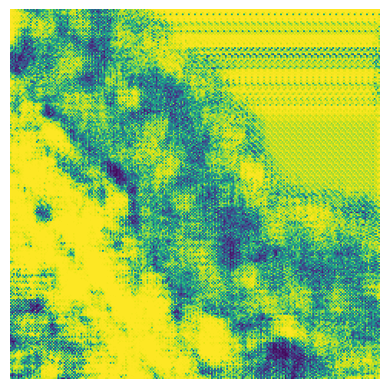} \\
\end{tabular}
\end{center}
\end{table*}

\begin{table*}[ht]
\caption{Detection performance comparison with raw images from the UW
dataset for three benchmarking methods.  The boldface and the underbar
indicate the best and the second-best results, respectively.}\label{tab:ori}
\begin{center}
\begin{tabular}{c|c|ccc}\hline\hline
Method & Features or Designs & F1 score  & Precision &  Recall \\ \hline
\multirow{8}{*}{Zhao, et al. (2021)} & Spatial  & 75.81\% & 78.15\%  & 73.61\%  \\
                       & Histogram & 78.99\% & 72.93\%  & 86.16\%  \\
                       & Frequency & 65.84\% & 49.07\%  & \textbf{100\%} \\
                       & Spatial + Histogram & 86.77\% & 82.78\%  & 91.17\% \\
                       & Spatial + Frequency & 77.02\% & 78.75\%  & 75.36\%  \\
                       & Histogram + Frequency  & 83.90\% & 78.36\%  & 90.29\% \\
                       & Spatial + Histogram + Frequency  & 87.08\% & 82.73\% & 91.92\% \\\hline
\multirow{1}{*} DefakeHop (2021)  & & 96.89\% & 97.26\%  & 96.53\%  \\ \hline
\multirow{4}{*}{Geo-DefakeHop (Ours)}  & PixelHop A & \underline{99.88\%} 
                       & \textbf{100\%} & \underline{99.75\%}  \\
                       & PixelHop B & \textbf{100\%} 
                       & \textbf{100\%}  & \textbf{100\% } \\
                       & PixelHop C & \underline{99.88\%} 
                       & \textbf{100\%}  & \underline{99.75\%}  \\
                       & PixelHops A\&B\&C & \textbf{100\%} & \textbf{100\%}& \textbf{100\%}\\\hline
\end{tabular}
\end{center}
\end{table*}

We use an example to explain discriminant channel selection. It is a
PixelHop of dimension $3\times3\times3$, which has 27 channels in total.
The x-axis of Fig. \ref{fig:cp} is the channel index and the y-axis is
the energy percentage curve or the performance curve measured by the F1 score, where the F1 score 
will be defined in Sec. \ref{subsec:exp_setting}.  In these plots, blue lines
indicate that energy percentage of each channel while red, magenta and
green lines represent the F1 scores of the train, validation and
test data. We consider the following four settings.
\begin{enumerate}
\item \textbf{\textit{Raw images}} \\
A higher frequency channel usually has a higher performance score as
shown in Fig. \ref{fig:ori}.  Low-frequency channels are not as
discriminant as high-frequency channels. It validates our assumption
that the GANs fail to generate high-frequency components with high
fidelity. 
\item \textbf{\textit{Image resizing}} \\
The input image is resized from $256 \times 256$ to $64\times64$.  As
compared with the setting of raw images, the discriminant power of
high-frequency channels degrades a little bit as shown in Fig.
\ref{fig:resize}. This is attributed to the fact that the down-sampling
operation uses a low pass filter to alleviate aliasing. Despite the 
performance drop of each channel, the overall detection performance can 
be preserved by selecting more channels. 
\item \textbf{\textit{Additive Gaussian noise}} \\
Noisy satellite images are obtained by adding white Gaussian noise with
$\sigma = 0.1$, where the dynamic range of the input pixel is [0, 1].
Thus, the relative noise level is high. We see from Fig. \ref{fig:noise}
that low-frequency channels perform better than high-frequency channels.
This is because we need to take the signal-to-noise ratio (SNR) into
account. Low-frequency channels have higher SNR values than
high-frequency ones. As a result, low-frequency channels have higher
discriminant power. 
\item \textbf{\textit{JPEG compression}} \\
The experimental results with JPEG compression of quality factor 75 are
shown in Fig. \ref{fig:jpeg}. We see from the figure that the
performance of different channels fluctuates.  Generally, the
performance of low-frequency channels is better than that of
high-frequency channels since the responses of high-frequency channels
degrade due to higher quantization errors in JPEG compression. However,
We still can get discriminant channels based on the performance of the 
validation data. 
\end{enumerate}
Generally, if only one PixelHop is used, we select several most
discriminant channels for ensembles. If multiple PixelHops are used
simultaneously, we select one to two most discriminant channels from
each PixelHop for ensembles. All selections are based on the F1 score
performance of the validation dataset. 

\subsection{Image-level decision ensemble}

In the last stage, we ensemble predicted scores of all blocks in one
image tile. Let $N_{ch}$ denote the total number of selected channels.
Since each channel has one predicted score from the previous step, each
block has a feature vector of dimension $N_{ch}$.  For each image, we
concatenate the feature vectors of all blocks to form one feature vector
of the image.  Since there are 256 blocks in one tile. The dimension of
the image-level feature vector $256 N_{ch}$. An XGBoost classifier is
trained to determine the final prediction of each tile. $N_{ch}$ is a
hyperparameter that is decided by the performance of the validation
dataset. 

\subsection{Visualization of Detection Results}

An attacker may stitch real/fake image blocks to form an image tile so
as to confuse the ensemble classifier trained above. This can be
addressed by providing a visualization tool to show the detection result
in a local region. One example is given in Table \ref{tab:example}.  The
table has four columns. The first two columns show four real images and
their associated heat maps while the last two columns show four fake
images and their associated heat maps. The color of local region in heat
maps indicates the probability of being real or fake in that region.
Cold and warm colors mean a higher probability of being real and fake,
respectively. The heat map can be generated by averaging prediction
scores of overlapping blocks with a smaller stride.  To gain more
insights, we show channel-wise Saab features and channel-wise heat map
for the DC, AC1, AC11 and AC26 frequencies of a PixelHop of dimension $3
\times 3 \times 3$ in Table \ref{tab:features}, where DC and AC1 are
low-frequency channels, AC11 is a mid-frequency channel and AC26 is a
high-frequency channel.  As shown in the last column of the table, DC
and AC1 are not as discriminant as AC11 and AC26. 

\begin{table*}[ht]
\caption{Detection performance comparison for images resized from $256
\times 256$ to $128 \times 128$ and $64 \times 64$ The boldface and the
underbar indicate the best and the second-best results, respectively.}\label{tab:resize}
\begin{center}
\begin{tabular}{c|c|c|ccc}\hline\hline
Tile size & Method & Features or Designs & F1 score  & Precision &  Recall \\ \hline
\multirow{11}{*}{128 x 128} &\multirow{8}{*}{Zhao, et al. (2021)} & Spatial & 77.35\% & 76.61\%  & 78.10\%  \\
                       & & Histogram & 80.09\% & 75.93\%  & 84.72\%  \\
                       & & Frequency & 64.14\% & 47.21\%  & \textbf{100\%} \\
                       & & Spatial + Histogram & 88.28\% & 85.81\%  & 90.89\% \\
                       & & Spatial + Frequency & 79.79\% & 81.38\%  & 78.26\%  \\
                       & & Histogram + Frequency  & 81.92\% & 76.99\%  & 87.53\% \\
                       & & Spatial + Histogram + Frequency  & 88.09\% & 86.52\% & 89.71\% \\\cline{2-6}
&\multirow{1}{*}{DefakeHop (2021)}  &   & 86.60\% & 89.36\%  & 84.00\%  \\\cline{2-6}
&\multirow{4}{*}{Geo-DefakeHop (Ours)}  & PixelHop A & \textbf{100\%} & \textbf{100\%}  & \textbf{100\%}  \\
                      & & PixelHop B & \underline{99.88\%} & \textbf{100\%} & \underline{99.75\%}  \\
                      & & PixelHop C & 99.75\% & \underline{99.75\%} & \underline{99.75}\%  \\
                      & & PixelHops A\&B\&C  & \textbf{100\%} & \textbf{100\%}  & \textbf{100\%}  \\\hline
\hline\hline

\multirow{11}{*}{64 x 64}  &\multirow{8}{*}{Zhao, et al. (2021)} & Spatial & 76.46\% & 78.85\%  & 74.21\%  \\
                       & & Histogram & 81.59\% & 76.60\%  & 87.26\%  \\
                       & & Frequency & 49.75\% & 79.89\%  & 36.12\% \\
                       & & Spatial + Histogram & 88.22\% & 86.15\%  & 90.39\% \\
                       & & Spatial + Frequency & 77.46\% & 77.83\%  & 77.09\%  \\
                       & & Histogram + Frequency  & 83.16\% & 77.80\%  & 89.32\% \\
                       & & Spatial + Histogram + Frequency  & 87.91\% & 83.94\% & 92.29\% \\\cline{2-6}
&\multirow{1}{*}{DefakeHop (2021)}  &     & 92.63\% & 97.78\%  & 88.00\%  \\\cline{2-6}
&\multirow{4}{*}{Geo-DefakeHop (Ours)}  & PixelHop A & \underline{98.27\%} & \underline{98.27\%}  & \underline{98.27\%}  \\
                      & & PixelHop B & 97.39\% & 97.76\%  & 97.03\%  \\
                      & & PixelHop C & 96.36\% & 97.71\%  & 95.05\%  \\
                      & & PixelHops A\&B\&C & \textbf{99.01\%} &\textbf{ 99.01\%}  & \textbf{99.01\% } \\\hline
                       
\end{tabular}
\end{center}
\end{table*}

\section{Experiments}\label{sec:experiments}

\subsection{Dataset}

The UW Fake Satellite Image dataset \citep{zhao2021deep} is used to
evaluate the proposed Geo-DefakeHop method. This is the first publicly
available dataset targeting at authentic and fake satellite images
detection.  Its authentic satellite images are collected from Google
Earth’s satellite images while its fake satellite images are generated
by CycleGAN. The base map used to generate fake satellite images are
from CartoDB \citep{cartodb2021}.  There are 4032 authentic color satellite images
of spatial resolution $256 \times 256$ and their fake counterparts in
the dataset. It covers Tacoma, Seattle and Beijing three cities.

\subsection{Experiment settings}\label{subsec:exp_setting}

We compare the performance of three methods: 1) the method proposed by
Zhao {\em et al.} \citep{zhao2021deep}, 2) DefakeHop
\citep{chen2021defakehop}, and 3) Geo-Defakehop.  We follow the same
experimental setting as given in \citep{zhao2021deep}. The dataset is
randomly split into training (90\%) and test sets (10\%). The model is
obtained by the training set and evaluated on the test set. In order to
fine-tune hyperparameters, we further split 90\% training images into
two parts: 80\% for model training and 10\% for the validation purpose. 
For Geo-DefakeHop, we consider four PixelHop designs:
\begin{itemize}
\itemsep -1ex
\item PixelHop A: Selected discriminant channels from 12 filters of dimension $2\times2\times3$, 
\item PixelHop B: Selected discriminant channels from 27 filters of dimension $3\times3\times3$, 
\item PixelHop C: Selected discriminant channels from 48 filters of dimension $4\times4\times3$,
\item PixelHop A\&B\&C: Selected discriminant channels from PixelHops A, B and C. 
\end{itemize}
We compare the detection performance under four settings:
\begin{itemize}
\itemsep -1ex
\item Raw images obtained from the UW dataset;
\item Image tiles being resized from $256\times 256$ to $128 \times 128$
and to $64 \times 64$;
\item Image tiles corrupted additive white Gaussian noise with standard
deviation $\sigma=0.02, 0.06, 0.1$;
\item Image tiles coded by the JPEG compression standard.
\end{itemize}
Training and test images go through the same image manipulation conditions.
As to the performance metrics, we use the F1 score, precision and recall as 
defined by
\begin{align*}
  \mbox{F1 score} &= 2 \times \frac{\mbox{precision} \times \mbox{recall}}{\mbox{precision} + \mbox{recall}},\\
  \mbox{precision} &= \frac{\mbox{true positive}}{\mbox{true positive}+\mbox{false positive}},\\
  \mbox{recall} &= \frac{\mbox{true positive}}{\mbox{true positive}+\mbox{false negative}}.
\end{align*}

\begin{table*}[ht]
\caption{Detection performance comparison for images corrupted by additive white Gaussian noise 
with standard deviation $\sigma=0.02, 0.06, 0.1$. The boldface and the underbar indicate the
best and the second-best results, respectively.}\label{tab:noise}
\begin{center}
\begin{tabular}{c|c|c|ccc}\hline\hline
Noise $\sigma$ & Method & Features or Designs & F1 score  & Precision &  Recall \\ \hline
\multirow{11}{*}{0.02} &\multirow{8}{*}{Zhao, et al. (2021)} & Spatial & 70.74\% & 72.58\%  & 68.98\%  \\
                       & & Histogram & 81.24\% & 75.00\%  & 88.60\%  \\
                       & & Frequency & 44.52\% & 67.66\%  & 33.17\% \\
                       & & Spatial + Histogram & 83.04\% & 82.41\%  & 83.67\% \\
                       & & Spatial + Frequency & 75.63\% & 78.42\%  & 73.04\%  \\
                       & & Histogram + Frequency  & 81.47\% & 76.62\%  & 86.98\% \\
                       & & Spatial + Histogram + Frequency  & 83.25\% & 81.47\% & 85.11\% \\\cline{2-6}

&\multirow{1}{*}{DefakeHop (2021)}  &   & 91.84\% & 93.75\%  & 90.00\%  \\\cline{2-6}
&\multirow{4}{*}{Geo-DefakeHop (Ours)}  & PixelHop A & 97.56\% & 96.38\%  & 98.77\%  \\
                      & & PixelHop B & \underline{98.90\%} & \underline{98.05\%}  & \textbf{99.75\%}  \\
                      & & PixelHop C & \textbf{99.01\%} & \textbf{98.53\%}  & \underline{99.50\%}  \\
                      & & PixelHop A\&B\&C & 98.65\% & 97.58\%  & \textbf{99.75\%}  \\\hline

\hline\hline

\multirow{11}{*}{0.06}  &\multirow{8}{*}{Zhao, et al. (2021)} & Spatial & 68.22\% & 74.77\%  & 62.72\%  \\
                       & & Histogram & 78.53\% & 71.70\%  & 86.80\%  \\
                       & & Frequency & 65.39\% & 48.57\%  & 100\% \\
                       & & Spatial + Histogram & 80.74\% & 80.94\%  & 80.54\% \\
                       & & Spatial + Frequency & 76.39\% & 78.47\%  & 74.42\%  \\
                       & & Histogram + Frequency  & 80.28\% & 75.49\%  & 85.71\% \\
                       & & Spatial + Histogram + Frequency  & 81.42\% & 79.40\% & 83.55\% \\\cline{2-6}
&\multirow{1}{*}{DefakeHop (2021)}  &           & 92.78\% & 95.75\%  & 90.00\%  \\\cline{2-6}
&\multirow{4}{*}{Geo-DefakeHop (Ours)}  & PixelHop A & \underline{95.24\%} & 93.98\%  & 96.53\%  \\
                      & & PixelHop B & \textbf{96.59\%} & \textbf{95.19\%}  & \textbf{98.02\% } \\
                      & & PixelHop C & 95.07\% & \underline{94.70\%}  & \underline{97.28\%}  \\
                      & & PixelHop A\&B\&C & \textbf{96.59\%} & \textbf{95.19\%}  & \textbf{98.02\%}  \\\hline
\hline\hline

\multirow{11}{*}{0.1}  &\multirow{8}{*}{Zhao, et al. (2021)} & Spatial  & 68.58\% & 69.76\%  & 67.44\%  \\
                       & & Histogram & 77.13\% & 71.37\%  & 83.91\%  \\
                       & & Frequency & 65.92\% & 78.48\%  & 56.83\% \\
                       & & Spatial + Histogram & 81.74\% & 78.42\%  & 85.35\% \\
                       & & Spatial + Frequency & 69.05\% & 70.35\%  & 67.79\%  \\
                       & & Histogram + Frequency  & 79.44\% & 74.67\%  & 84.86\% \\
                       & & Spatial + Histogram + Frequency  & 80.05\% & 77.78\% & 82.46\% \\\cline{2-6}
&\multirow{1}{*}{DefakeHop (2021)}  &   & 92.63\% & 97.78\%  & 88.00\%  \\\cline{2-6}
&\multirow{4}{*}{Geo-DefakeHop (Ours)}  & PixelHop A & 94.43\% & 92.42\%  & 96.53\%  \\
                      & & PixelHop B & 94.88\% & 93.51\%  & 96.29\%  \\
                      & & PixelHop C & \underline{95.37\%} & \underline{93.99\%}  & \underline{96.78\%} \\
                      & & PixelHop A\&B\&C & \textbf{96.10\%} & \textbf{94.71\%}  & \textbf{97.52\%}  \\\hline
                       
\end{tabular}
\end{center}
\end{table*}

\subsection{Detection performance comparison}\label{subsec:performance}

We compare the performance of three detection methods under various
conditions in this subsection. 

{\bf Raw images.} We conduct both training and testing on raw images
from the UW dataset \citep{zhao2021deep} and show the performance of the
three methods in Table \ref{tab:ori}.  As shown in the table, we see
that PixelHop B and PixelHop A\&B\&C of Geo-DefakeHop achieve perfect
detection performance with 100\% F1 score, 100\% precision and 100\%
recall while PixelHop A and PixelHop B achieve nearly perfect
performance. Both Geo-DefakeHop and DefakeHop outperform Zhao {\em et
al.}'s method in all performance metrics by significant margins. There
is also a clear performance gap between Geo-DefakeHop and DefakeHop. 

{\bf Image resizing.} Image resizing is a common image manipulation
operation. For the image tile of resolution $256 \times 256$ in the UW
dataset, its scale with respect to the physical size is 1:8000. We
resize the tile from $256 \times 256$ to images of lower resolutions
(i.e.  $128 \times 128$ and $64 \times 64$) and test the capability of
the three methods in authentic/fake image detection.  The results are
shown in Table \ref{tab:resize}. For image resized to $128 \times 128$,
both PixelHop A and PixelHop A\&B\&C achieve perfect performance with
100\% F1 score, 100\% precision and 100\% recall while PixelHop B and
PixelHop C achieve nearly perfect performance. For image resized to $64
\times 64$, we see the power of ensembles. That is, the F1 score,
precision and recall of PixelHop A\&B\&C are all above 99\%, which is
slightly better than an individual PixelHop. Again, all four
Geo-DefakeHop settings outperform Zhao {\em et al.}'s method by
significant margins.  DefakeHop is slightly better than Zhao {\em et
al.}'s method but significantly worse than Geo-DefakeHop. 

\begin{table*}[ht]
\caption{Detection performance comparison for images coded by the JPEG
compression standard of three quality factors (QF), i.e., $QF=95$, $85$
and $75$.  The boldface and the underbar indicate the best and the
second-best results, respectively.}\label{tab:jpeg}
\begin{center}
\begin{tabular}{c|c|c|ccc}\hline\hline

JPEG quality factor & Method & Features or Designs & F1 score  & Precision &  Recall \\ \hline
\multirow{11}{*}{95}  &\multirow{8}{*}{Zhao, et al. (2021)} & Spatial  & 74.88\% & 73.96\%  & 75.82\%  \\
                       & & Histogram & 76.05\% & 68.27\%  & 85.83\%  \\
                       & & Frequency & 64.37\% & 47.46\%  & 100\% \\
                       & & Spatial + Histogram & 85.95\% & 82.49\%  & 89.72\% \\
                       & & Spatial + Frequency & 78.00\% & 78.38\%  & 77.62\%  \\
                       & & Histogram + Frequency  & 82.43\% & 74.95\%  & 91.58\% \\
                       & & Spatial + Histogram + Frequency  & 86.96\% & 85.06\% & 88.94\% \\\cline{2-6}
&\multirow{1}{*}{DefakeHop (2021)}  &          & 98.00\% & 98.00\%  & 98.00\%  \\\cline{2-6}
&\multirow{4}{*}{Geo-DefakeHop (Ours)}  & PixelHop A & 97.91\% & 97.31\%  & 98.51\%  \\
                      & & PixelHop B & 97.90\% & 97.54\%  & 98.27\%  \\
                      & & PixelHop C & \textbf{ 98.28\%} & \textbf{97.56\%}  & \textbf{99.01\%}  \\
                      & & PixelHop A\&B\&C & \underline{98.15\%} & \underline{97.55\%}  & \underline{98.76\%}  \\\hline
\hline\hline

\multirow{11}{*}{85} &\multirow{8}{*}{Zhao, et al. (2021)} & Spatial  & 76.46\% & 78.64\%  & 74.38\%  \\
                       & & Histogram & 77.99\% & 72.71\%  & 84.09\%  \\
                       & & Frequency & 78.05\% & 75.85\%  & 80.38\% \\
                       & & Spatial + Histogram & 85.91\% & 82.67\%  & 89.42\% \\
                       & & Spatial + Frequency & 82.53\% & 81.48\%  & 83.61\%  \\
                       & & Histogram + Frequency  & 85.28\% & 81.66\%  & 89.24\% \\
                       & & Spatial + Histogram + Frequency  & 89.54\% & 85.82\% & 93.6\% \\\cline{2-6}
&\multirow{1}{*}{DefakeHop (2021)}  &   & 92.93\% & 93.88\%  & 92.00\%  \\\cline{2-6}
&\multirow{4}{*}{Geo-DefakeHop (Ours)}  & PixelHop A & \underline{97.54\%} & \underline{96.83\%}  &\underline{ 98.27\%}  \\
                      & & PixelHop B & \textbf{97.91\%} & \textbf{97.08\%}  & \textbf{98.76\%}  \\
                      & & PixelHop C & \textbf{97.91\%} & \textbf{97.08\%}  & \textbf{98.76\%}  \\
                      & & PixelHop A\&B\&C & \underline{97.54\%} & 97.06\%  & 98.02\%  \\\hline
\hline\hline

\multirow{11}{*}{75}  &\multirow{8}{*}{Zhao, et al. (2021)} & Spatial  & 73.88\% & 75.78\%  & 72.03\%  \\
                       & & Histogram & 80.96\% & 76.70\%  & 85.71\%  \\
                       & & Frequency & 80.77\% & 79.15\%  & 82.47\% \\
                       & & Spatial + Histogram & 85.61\% & 81.70\%  & 89.93\% \\
                       & & Spatial + Frequency & 87.09\% & 83.94\%  & 90.49\%  \\
                       & & Histogram + Frequency  & 88.94\% & 87.41\%  & 90.52\% \\
                       & & Spatial + Histogram + Frequency  & 90.20\% & 88.46\% & 92.00\% \\\cline{2-6}
&\multirow{1}{*}{DefakeHop (2021)}  &     & 94.85\% & 97.87\%  & 92.00\%  \\\cline{2-6}
&\multirow{4}{*}{Geo-DefakeHop (Ours)}  & PixelHop A & \textbf{97.92\%} & 96.63\%  & \textbf{99.26\%}  \\
                      & & PixelHop B & 97.66\% & \textbf{97.07\%}  & 98.27\%  \\
                      & & PixelHop C & \underline{97.79\%} & \underline{96.84\%}  & \underline{98.76\%}  \\
                      & & PixelHop A\&B\&C & \textbf{97.92\%} & 96.63\%  & \textbf{99.26\% } \\\hline
                       
\end{tabular}
\end{center}
\end{table*}

{\bf Additive white Gaussian noise.} Satellite images can be corrupted
by additive white Gaussian noise due to long distance image transmission
or image acquisition. By normalizing the image pixel value to the range
of $[0,1]$, we test the detection performance with three noise levels
$\sigma=0.02, 0.06, 0.1$ and show the results in Table \ref{tab:noise}.
We see from the table that, if authentic or fake satellite images are
corrupted by white Gaussian noise with $\sigma=0.02$, $0.06$ and $0.1$,
the F1 scores of Geo-DefakeHop decreases from 100\% to 99.01\%, 96.59\%
and 96.10\%, respectively. In contrast, the F1 scores of DefakeHop are
slightly above 90\% and those of Zhao {\em et al.}'s method are around
80\% or lower. Also, the ensemble gain of multiple PixelHops is more
obvious as the noise level becomes higher. 

{\bf JPEG compression.} JPEG compression is commonly used for image
sharing over the Internet. JPEG is a lossy compression method, where
higher spatial frequencies are quantized by larger quantization
step sizes. One can adjust the quality factor (QF) to get different
trade offs between quality and coding bit rates. The QF value is between
0 and 1. The image has the best quality but the highest bit rate when QF
is equal to 1. Typically, QF is chosen from the range of [0.7,1].  In
this experiment, we encode satellite images by JPEG with QF=0.95, 0.85
and 0.75 and investigate the robustness of benchmarking methods against
these QF values. The results are shown in Table \ref{tab:jpeg}.  The F1
scores of Geo-DefakeHop are 98.28\%, 97.91\% and 97.92\% for QF=0.95,
0.85 and 0.75, respectively. 

By comparing the three distortion types, the additive white Gaussian
noise has the most negative impact on the detection performance, JPEG
compression the second, and image resizing has the least impact. This is
consistent with our intuition. Image resizing does not change the
underlying information of images much, JPEG changes the information
slightly because of the fidelity loss of high-frequencies and the
additive white Gaussian noise perturbs the information of all
frequencies. 

\subsection{Model size computation}

For a PixelHop of filter size $s1 \times s2\times c$, it has at most
$M_{\max} = s1 \times s2 \times c$ filters.  For example, the size of
PixelHop A is $2 \times 2 \times 3$ and $M_{A,\max}=12$. Similarly, we
have $M_{B,\max}=27$ and $M_{C,\max}=48$.  However, we choose only a
subset of discriminant filters. They are denoted by $M_A$, $M_B$ and
$M_C$, respectively. 

\begin{table*}[ht]
\vspace{-0.3cm}
\caption{Model size computation of four Geo-DefakeHop designs for raw satellite 
input images.}\label{tab:parameters}
\begin{center}
\vspace{0.1cm}
\begin{tabular}{cccccc} \hline\hline
System & No. of Selected & No. of Filter & No. of c/w XGBoost & No. of ensemble    & Total \\ 
System & Channels        & Parameters    &  Parameters        & XGBoost Parameters & Model Size\\ \hline
Pixelhop A & 1       & 12   & 400            & 400                & 812 \\ 
Pixelhop B & 1       & 27   & 400            & 400                & 827 \\ 
Pixelhop C & 1       & 48   & 400            & 400                & 848 \\ 
Pixelhop A\&B\&C & 3 & 87   & 1,200          & 1,200              & 2,487 \\ \hline
\end{tabular}
\end{center}
\vspace{-0.2cm}
\end{table*}

\begin{table*}[ht]
\caption{Summary of model sizes of four Geo-DefakeHop designs with different input 
images.}\label{tab:all_parameters}
\begin{center}
\begin{tabular}{ccccc}\hline\hline
 Experiments  & PixelHop A & PixelHop B & PixelHop C &  PixelHop A\&B\&C\\ \hline
 Raw Images   & 0.8K  & 0.8K  & 0.8K   & 2.5K  \\ 
 Resizing     & 9.7K  & 20K   & 37K    & 61.7K  \\
 Noise        & 8.1K  & 13K   & 33K    & 38.5K \\
 Compression  & 7.3K  & 19K   & 33K    & 37.4K  \\ \hline
\end{tabular}
\end{center}
\vspace{-0.2cm}
\end{table*}

An XGBoost classifier consists of a sequence of binary decision trees,
which are specified by two hyper-parameters: the max depth and the
number of trees.  Each XGBoost tree consists of both leaf nodes and
non-leaf nodes. Non-leaf nodes have two parameters (i.e., the dimension
and the value) to split the dataset where leaf nodes have one parameter
(i.e., the predicted value). We have two types of XGBoost classifiers: 1)
the channel-wise classifier and 2) the ensemble classifier. For the
former, the max depth and the number of the trees are set to 1 and 100 respectively. 
Since each tree has one non-leaf node and two leaf nodes, its model size
is $4 \times 100 =400$ parameters.  For the latter, the max
depth and the number of trees are set to 1 and $100 \times M$, where
$M=M_A+M_B+M_C$ is the total number of selected discriminant channels of
all three PixelHops, respectively. The model size of the ensemble
classifier is $4\times 100 \times M=400M$ parameters.

As an example, we provide the model size computation detail in Table
\ref{tab:parameters} for four Geo-DefakeHop designs with the raw
satellite images as the input.  As shown in the table, PixelHops A, B, C
and A\&B\&C have 812, 827, 848 and 2,487 parameters, respectively.
Since the selected discriminant channel numbers of PixelHops A, B, C and
A\&B\&C vary with raw, resized, noisy and compressed input satellite
images, their model sizes are different. The model sizes are
summarized in Table \ref{tab:all_parameters}. 

\section{Conclusion and Future Work}\label{sec:conclusion}

A method called Geo-DefakeHop was proposed to distinguish between authentic and
counterfeit satellite images. The design may contain a single PixelHop
of different filter sizes or the ensemble of discriminant channels of
multiple PixelHops. For the former, three design choices called PixelHop
A, PixelHop B and PixelHop C were tested. For the latter, we considered
a design by ensembles of discriminant channels of PixelHop A, PixelHop B
and PixelHop C. The channel-wise performance analysis offers
interpretability. The effectiveness of the Geo-DefakeHop method in terms
of the F1 scores, precision and recall was demonstrated by extensive
experiments. Furthermore, the model sizes of the four designs were
thoroughly analyzed. They can be easily implemented in software on
mobile or edge devices due to small model sizes. 

The UW Fake Satellite Image dataset only contains Tacoma, Seattle and
Beijing three cities. A large-scale fake satellite image dataset with
more cities can be constructed to make the dataset more
challenging. Besides, the current dataset was built by one GAN.  One may
build more fake satellite images with multiple GANs as well as computer
graphics techniques. Furthermore, more manipulations such as blurring
and contrast adjustment can be added to test the limitation of the
detection system. 

\section{Acknowledgments}\label{sec:acknowledgments}

This work was supported by the Army Research Laboratory (ARL) under
agreement W911NF2020157. 

\printcredits
\bibliographystyle{cas-model2-names}
\bibliography{refs}
\end{document}